\documentclass{article}
\usepackage[final]{neurips_2021}

\usepackage[utf8]{inputenc} 
\usepackage[T1]{fontenc}    
\usepackage{hyperref}       
\usepackage{url}            
\usepackage{booktabs}       
\usepackage{amsfonts}       
\usepackage{nicefrac}       
\usepackage{microtype}      
\usepackage{xcolor}         
\usepackage{graphicx}
\graphicspath{{images/}}  
\usepackage{comment} 
\usepackage{array}

\title{Designing A Clinically Applicable Deep Recurrent Model to Identify Neuropsychiatric Symptoms in People Living with Dementia Using In-Home Monitoring Data}
\author{%
        Francesca Palermo\thanks{The authors are part of the Care Research and Technology Centre, The UK Dementia Research Institute} \\
        Department of Brain Sciences\\
        Imperial College of London\\
        London, UK \\
        \texttt{f.palermo@imperial.ac.uk} \\
    \And
        Honglin Li$^{*}$ \\
        Department of Brain Sciences\\
        Imperial College of London\\
        London, UK \\
        \texttt{honglin.li20@imperial.ac.uk} \\
    \And
        Alexander Capstick$^{*}$\\
        Department of Brain Sciences\\
        Imperial College of London\\
        London, UK \\
        \texttt{alexander.capstick19@imperial.ac.uk} \\
    \And
        Nan Fletcher-Lloyd$^{*}$\\
        Department of Brain Sciences\\
        Imperial College of London\\
        London, UK \\
        \texttt{nan.fletcher-lloyd17@imperial.ac.uk} \\
    \And
        Yuchen Zhao$^{*}$\\
        Dyson School of Design Engineering\\
        Imperial College London\\
        London, UK\\
        \texttt{yuchen.zhao19@imperial.ac.uk}\\
    \And
        Samaneh Kouchaki$^{*}$\\
        Centre for Vision, Speech, and Signal Processing\\
        University of Surrey\\
        Guildford, UK \\
        \texttt{s.kouchaki@imperial.ac.uk}\\
    \And
        Ramin Nilforooshan$^{*}$\\
        Surrey and Borders Partnership  \\ 
        NHS Foundation  Trust\\
        Surrey, UK \\
        \texttt{ramin.nilforooshan@sabp.nhs.uk}\\
    \And
        David Sharp$^{*}$\\
        Department of Brain Sciences\\
        Imperial College of London\\
        London, UK \\
        \texttt{david.sharp@imperial.ac.uk}\\
    \And
        Payam Barnaghi$^{*}$\\
        Department of Brain Sciences\\
        Imperial College of London\\
        London, UK \\
        \texttt{p.barnaghi@imperial.ac.uk} \\
}

\begin{document}

\maketitle

\begin{abstract}
    Agitation is one of the neuropsychiatric symptoms with high prevalence in dementia which can negatively impact the Activities of Daily Living (ADL) and the independence of individuals.
    Detecting agitation episodes can assist in providing People Living with Dementia (PLWD) with early and timely interventions.
    Analysing agitation episodes will also help identify modifiable factors such as ambient temperature and sleep as possible components causing agitation in an individual.
    This preliminary study presents a supervised learning model to analyse the risk of agitation in PLWD using in-home monitoring data. 
    The in-home monitoring data includes motion sensors, physiological measurements, and the use of kitchen appliances from 46 homes of PLWD between April 2019-June 2021.  
    We apply a recurrent deep learning model to identify agitation episodes validated and recorded by a clinical monitoring team. 
    We present the experiments to assess the efficacy of the proposed model.
    The proposed model achieves an average of 79.78\% recall, 27.66\% precision and 37.64\% F1 scores when employing the optimal parameters, suggesting a good ability to recognise agitation events.
    We also discuss using machine learning models for analysing the behavioural patterns using continuous monitoring data and explore clinical applicability and the choices between specificity and specificity in home monitoring applications.   
\end{abstract}



\section{Introduction}
There are currently 50 million People Living with Dementia (PLWD) around the world, with the number expected to triple in the next 30 years \cite{livingston2020dementia,patterson2018world}. 
While there are medications available to help control symptoms and delay the progression of dementia \cite{overshott2005treatment}, there is yet no pharmacological cure.
PLWD often rely on care and support from family members and/or healthcare professionals. 
Hence, the impact of dementia on caregivers, health and social care services is significant. 

In addition to symptoms associated with cognitive decline such as memory loss and problems with attention \cite{blankevoort2010review}, up to 90\% of PLWD develop behavioural and psychological symptoms of dementia (BPSD) during the progression of the disease. 
These symptoms may include sleep disturbances and changes in affective behaviours, such as apathy. 
Among these symptoms, agitation is the most frequent \cite{feast2016behavioural}, affecting at least 25\% of all PLWD \cite{bankole2011continuous}, and negatively impacting Activities of Daily Living (ADL).
Agitation can manifest itself in various ways, including repetitive behaviour, wandering, and socially inappropriate conduct, and can progress to aggressive and dangerous episodes.

The provision of timely and effective intervention is a significant challenge in dementia care and requires frequent and reliable monitoring of PLWD. 
Internet of Things (IoT) technologies such as wearable devices, connected vital sign monitoring devices and environmental and activity monitoring sensors can be used to monitor the day-to-day well-being of PLWD. 
These technologies could allow the retention of autonomy to PLWD and provide peace of mind to caregivers \cite{buchanan2010role}.
Moreover, the integration of machine learning and in-home monitoring devices has the potential to identify changes in physical and psychological well-being. 
Machine learning approaches can analyse changes in the activity and health patterns and allow preventative interventions to avoid or reduce the scale of the negative long-term impact. 

In this preliminary work, we present a supervised deep learning model to analyse the risk of agitation in PLWD. 
The model is applied to the data acquired from 46 homes of PLWD using physiological and in-home passive infrared (PIR) sensors in a study within the UK Dementia Research Institute Care Research and Technology Centre (UK DRI CR\&T). 

The paper is organised as follows.
Section \ref{sec:data_collection} introduces the dataset used in this work.
Section \ref{sec:related_work} presents a brief survey of the current state of the art on identifying neuropsychiatric symptoms in PLWD.
Section \ref{sec:methodology} describes the experimental methodology of the work, including preprocessing of the dataset and labels and design of the implemented model.
Section \ref{sec:experiments_setup} introduces the results of the proposed experiments which are further discussed in Section \ref{sec:discussion}.
Section \ref{sec:conclusion} presents the conclusion and future work.

\section{Collecting and Integrating Remote Monitoring Data}
\label{sec:data_collection}
\begin{figure}
    \centering
    \includegraphics{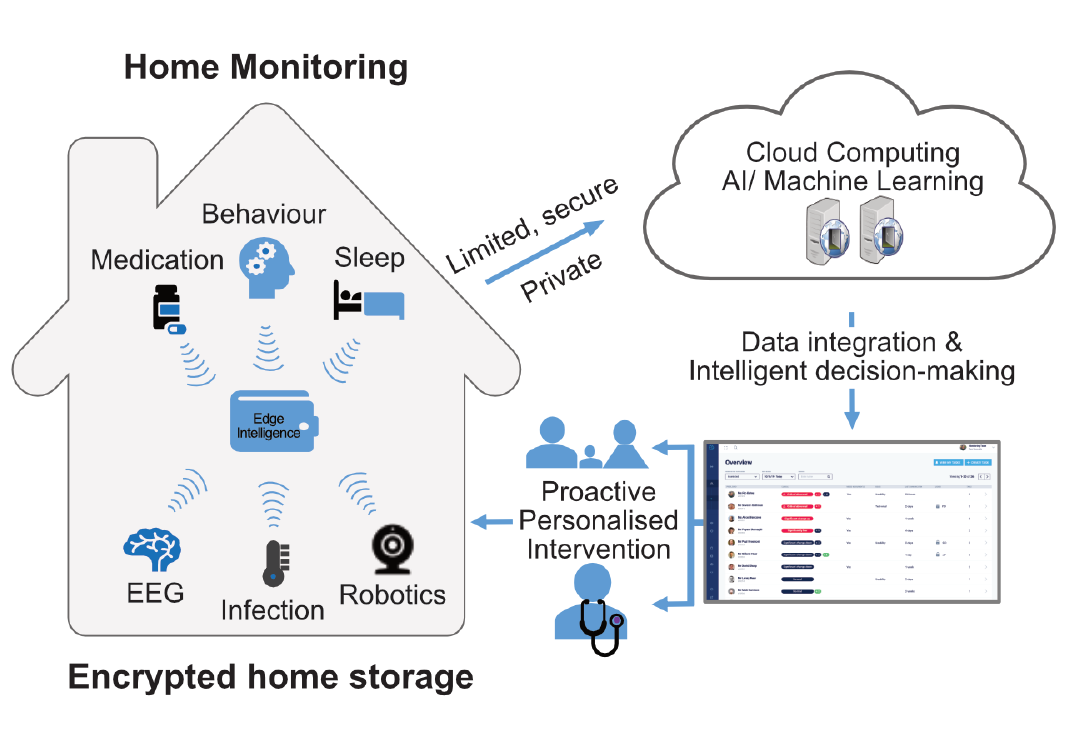}
    \caption{An overview of the in-home monitoring system. 
    Data is communicated from devices to an edge device and integrated into a secure Cloud. 
    A web-based interface is used to visualise the data. 
    A monitoring team view the data and respond to alerts and interact with carers and PLWD.}
    \label{fig:overview}
\end{figure}

The dataset used in this study is a subset of the data collected in a clinical study by the UK DRI CR\&T Centre to improve care for PLWD using technology. 
Almost $23\%$ of unplanned hospital admissions for PLWD are due to potentially preventable causes \cite{phe2015}. 
The data is collected using a variety of in-home monitoring devices that provide multi-modal environmental information, physiological and sleep markers.
A digital platform, called Minder is developed to integrate the data from heterogeneous sources and devices as well as analytical and predictive models. 
The platform provides an overview dashboard that allows a clinical monitoring team to see the raw observation and measurement data and also flags raised by analytical and predictive models. 
The monitoring team then follows a clinical protocol to respond to the observations and flags and contact PLWD, their carers or health and care services depending on the requirements. 
Figure \ref{fig:overview} provides an overview of the framework of the data collection and intervention study. 

A list of the devices and the frequency of data collection is shown in Table \ref{tab:sensor_description}.
\begin{table}[t]
    \centering
    \caption{Digital markers in in-home monitoring in dementia care}
    \begin{tabular}{>{\centering\arraybackslash}p{.3\linewidth} >{\centering\arraybackslash}p{.3\linewidth} >{\centering\arraybackslash}p{.3\linewidth}}
        \toprule
        \textbf{Digital marker}  & \textbf{Monitoring Device}  & \textbf{Frequency}
        \\
        \midrule
          Movement & Door, Motion and Tracking sensors  & Continuous (multiple doors, hallway, kitchen, and rooms)
        \\
        \midrule
         Home Device Usage & Home appliance use, smart plugs & Continuous (kitchen appliances, TV, and other commonly used devices) 
         \\   
        \midrule
         Body temperature & Smart Temporal Thermometer & Twice daily (morning, evening) or continuous using a wearable device 
        \\    
        \midrule
         Blood Pressure & BPM Connect or using a wearable device &  Twice a day (morning, evening) or frequent times daily 
         \\  
        \midrule
        Pulse & Pulse HR  & Continuous   
        \\
        \midrule
        Weight & Smart Scale with Body Composition \& Heart Rate  & Once a day 
        \\
    \end{tabular}
    \label{tab:sensor_description}
\end{table}

Mini–Mental State Examination (MMSE) tests were also given to PLWD to assess their executive functions. 
This examination consists of various questions designed to test a PLWD's everyday mental skills. 
It includes tests such as recalling names of common objects after a few minutes and drawing a clock with the numbers in the correct positions. 
The score can range between $0 - 30$, with a lower score representing a lower mental ability.

Table \ref{tb:Age_MMSE_Stats} summarises the age and MMSE scores of the analysed dataset of 46 PLWD.
Further information on the dataset are introduced in Section \ref{sec:experiments_setup}. 
The participants had an average age of $82.6$ and a standard deviation of $7.4$; with the ages between $61$ and $99$ years. 
The subjects' MMSE scores have a mean value of $24.1$ and a standard deviation of $3.5$; with the lowest score of $15$ and the highest value of $30$. 
Within our dataset, we see little relationship between age and MMSE with a coefficient of regression of $0.1$ and an $R^2$ value of $3\%$.
\begin{table}[h!]
    \centering
     \caption{Age and MMSE statistics for the study cohort}
     \begin{tabular}{c| c c c c} 
         \toprule
         \textbf{Feature} & \textbf{Mean} & \textbf{Standard Deviation} & \textbf{Min} & \textbf{Max}\\ 
         \midrule
         \textbf{Age} & 82.5 & 7.2 & 61 & 99 \\ 
         \textbf{MMSE} & 23.8 & 3.6 & 15 & 30 \\ 
     \end{tabular}
     \label{tb:Age_MMSE_Stats}
\end{table}
Table \ref{tb:dementia_daignosis_stats} shows the distribution of the dementia diagnosis for $72\%$ of the total PLWD in our cohort, with Alzheimer's Disease accounting for the vast majority of these diagnoses.
\begin{table}[h!]
    \centering
    \caption{Dementia Diagnosis for the study cohort}
     \begin{tabular}{c| c c c } 
         \toprule
         \textbf{Dementia Type} & \textbf{\%Male} & \textbf{\%Female} & \textbf{\%Total} \\ 
         \midrule
         \textbf{Alzheimer's Disease} & 39\% & 15\% & 54\% \\ 
         \textbf{Parkinson's Dementia}  & 2\% & 2\% & 4\%  \\ 
         \textbf{Fronto-temporal Dementia} & 2\% & 2\% & 4\%  \\
         \textbf{Vascular Dementia} & 4\% & 4\% & 9\% \\
         \textbf{Other/not specified} & 22\% & 7\% & 28\% \\
     \end{tabular}
     \label{tb:dementia_daignosis_stats}
\end{table}
\section{Identifying the Risk of Neuropsychiatric Symptoms Using Time-course Multi-modal Data}
\label{sec:related_work}
Continuous monitoring has previously shown potential in detecting and predicting possible episodes of agitation. 
A computer vision approach designed by~\cite{fook2007automated} used a camera mounted on the ceiling of a room to simulate a bed chamber in a hospital ward to recognise agitation behaviour in PLWD.
However, using video data and computer vision approaches raises significant privacy and security issues. 
Moreover, the complexity and variety of indoor location tracking scenarios hinder the scalability of vision-based solutions. 

Another research study suggested a combination of the Hidden-Markov model and support-vector machine to detect agitation states~\cite{fook2007automated}. 
Their work extracted features from the temporal segmentation maps of PLWD tracked via computer vision segmentation techniques to measure behavioural parameters. 
However, this study was performed in a laboratory environment and none of the subjects had an established diagnosis of dementia.

~\cite{gong2015home} used a multi-modal sensing platform to investigate the relationship between urinary incontinence and night-time agitation in people with Alzheimer's Disease.
This platform was made up of (i) two accelerometers positioned on the bed to record movement, (ii) two inertial measurement units to the left and right wrist of the participant to monitor sleep agitation, (iii) an acoustic sensor to record possible verbal agitation, and (iv) a sensor to detect incontinence on the bed. 
Privacy issues of recording sound and sensitivity and specificity of accelerometers in real-world and uncontrolled settings are key limiting issues for such approaches. 
Accelerometer data could be very noisy, and agitation will not only manifest itself at sleep time. 
~\cite{ghali1995temporal} used motion sensors positioned inside a shirt to investigate the connection between agitation and the time of the day. 
Their results showed that people in the early, middle, and advanced stages of Alzheimer's indicated signs of neuropsychiatric symptoms mainly before sunset, around sunset, and after sunset, respectively.

The risk of agitation and unusual behavioural patterns in PLWD was proposed by~\cite{enshaeifar2018health} using in-home sensory data. 
However, their proposed model showed issues with larger groups with various settings as their models did not generalise well to more diverse data.
The work in~\cite{rezvani2021semi} presented a semi-supervised model of activity patterns over 24 hours in eight different household locations aggregated within each hour of day to detect the likelihood of agitation in PLWD. 
However, this study was only based on movement sensors, which limited the specificity and sensitivity of the model. 
The research by~\cite{khan2019agitation,spasojevic2021pilot} presented a multi-modal sensor data analysis to detect agitation events in PLWD using Random Forest (RF) and support-vector machine. 
The participants had to wear an Empatica E4 wearable device that can extract motion and physiological indicators (blood volume pulse, electrodermal activity, and skin temperature). 
They found that their model achieved the best results when using both movement data and physiological data. 
However, using wearable devices that require frequent charging and provide noisy data limit the scalability and efficacy of their solution for longer-term use in large cohorts. 
In our study, we use mainly passive and low-cost sensors such as basic physiology and movement rather than video and sound to reduce the burden on participants and complexity and privacy issues. 
Moreover, we develop deep representation methods that can generalise to more diverse data. 
We then use the results of the analysis model to triage the risk of agitation. 
A monitoring team investigates the generated alerts and the associated data and contact PLWD or their carer to confirm the episodes and record the false positives. 
This human in the loop approach allows us to verify the machine-generated alerts and also gradually re-train and improve the model as more labelled data is collected.

\section{Methodology}
\label{sec:methodology}
This section describes the methodology implemented in this work to identify agitation episodes in PLWD using a deep recurrent model applied to continuous in-home monitoring data.
\subsection{Data pre-processing}
\label{sec:preprocessing}
\begin{figure}
    \centering
    \includegraphics{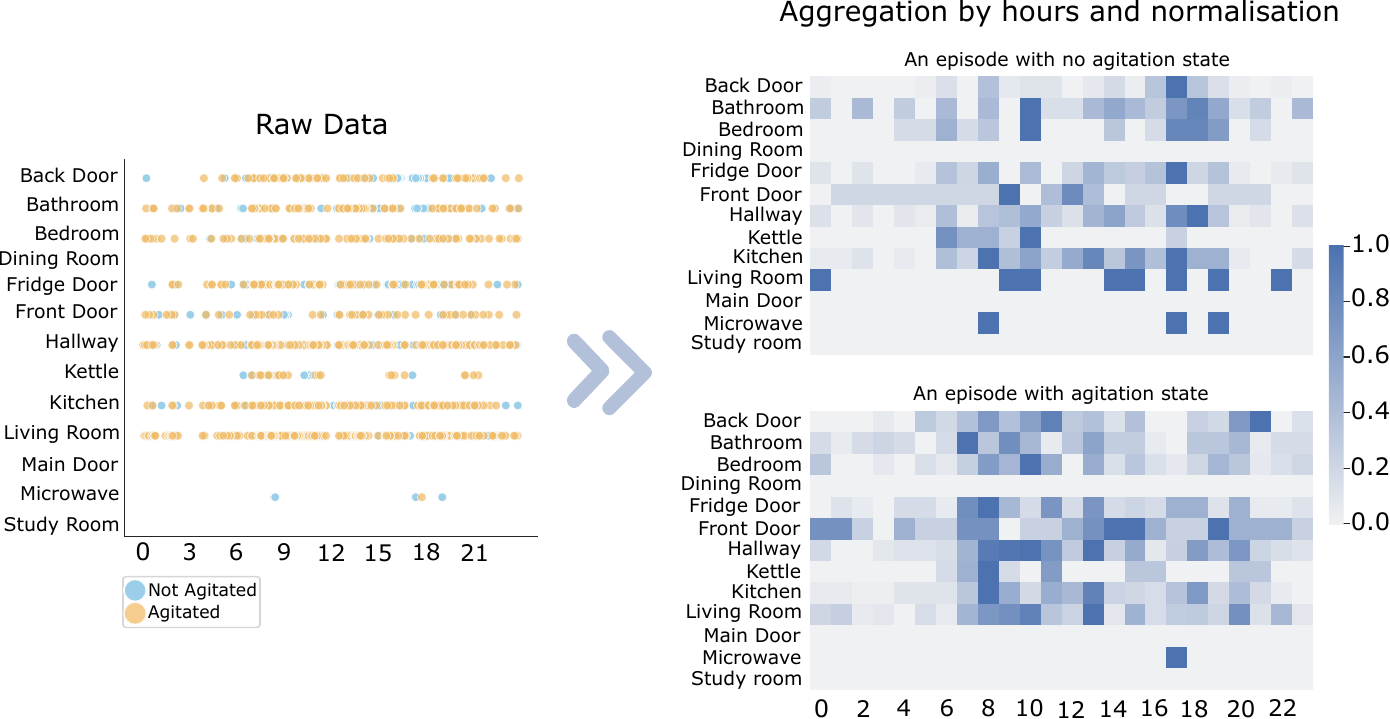}
    \caption{Visualisation of the movement data aggregated per hour and comparison of an individual's in-home movements in two different days, with and without an agitation episode. 
    The x-axis shows the time of the day (00-24) and the y-axis shows the location of the movement detection sensors in the home.}
    \label{fig:location_map}
\end{figure}

The data collected in this study come from heterogeneous sensor devices that collect data at different granularities. 
In order to harmonise various data modalities with different granularities, their temporal dimensions need to be aggregated.
First, the observations are grouped within each hour, and a list is constructed for hourly locations explored by each individual.
This provides an hour-by-hour map of the movements of the households.
The raw location data and aggregation and normalisation techniques are shown in Figure \ref{fig:location_map}, along with an individual's movements during a day with an episode of agitation versus a day without any agitation episodes for the same individual. 
Unusual patterns of activity could be an indication of an agitation episode. 
However, a simple inspection of the irregular patterns is insufficient to detect the agitation episodes and will not be clinically applicable due to low sensitivity and specificity. 
An effective analysis model needs to be constructed to learn the signature of patterns, identify the unusual patterns, and use previously seen examples to train a generalisable predictive model. 
The movement data from each in-door location for each hour is used as a feature to construct the input data into the model. 
The number of out of range pulse and blood pressure measurements are identified in the last 24 hours and included as additional features. 
The latter is due to the fact that changes in vital signs could also relate to agitation episodes (e.g., higher blood pressure and heartbeat). 
The minimum, maximum, and average blood pressure measurements and heart rate values in the past three days are also added to the input features.
Table \ref{tb:features} lists the features and their granularity used for the input data. 
We have only included data from participants that had a labelled and validated episode of agitation in the existing dataset. 
The input data includes data from the homes of 46 PLWD and has 24 features in total.
\begin{table*}[]
		\centering
		\caption{Set of features used in the experiments to identify the risk of agitation}
        \begin{tabular}{>{\centering\arraybackslash}p{.3\linewidth} >{\centering\arraybackslash}p{.3\linewidth} >{\centering\arraybackslash}p{.3\linewidth}}
            \toprule
            \textbf{Feature Type} & \textbf{Description} & \textbf{Granularity} \\ 
            \midrule
            Location/Home Device  & Motion and tracking sensors, as well as smart plugs, employed to track the participants' movements and appliance usage throughout the day: back door, bathroom, bedroom, dining room, fridge door, hallway, kitchen, living room, entrance door, microwave and study room  & Aggregated by the hour of the day  \\ 
            \midrule
            Physiological data  & Minimum, maximum and mean values of pulse and blood systolic and diastolic pressure & Aggregated values for the past three days   \\ 
            \midrule
            Out of range measurements alerts & Alerts generated based on out of range heart rate and blood pressure data & Number of alerts per day  \\ 
		\end{tabular}
		\label{tb:features}
\end{table*}

\subsection{Training Data}
The data is labelled as true (true positive) if the person has an episode of agitation validated by a clinical monitoring team, and false (true negative) if there was not an episode of agitation when the monitoring team contacted the PLWD. 
The initial labelling and verification were guided by our earlier work in this domain, which used an unsupervised model by applying Non-negative matrix factorisation for clustering the in-home activity data~\cite{enshaeifar2018health}. 
However, the early clustering model was only designed to guide the monitoring team to identify episodes and to create a set of labelled data. 
For clinically applicable solutions with acceptable sensitivity and specificity, we have been investigating generalisable models that can be used in a broader range of settings. 
In the current work, we have included physiological monitoring data to improve the efficacy of the model.  

\begin{figure}
    \centering
    \includegraphics{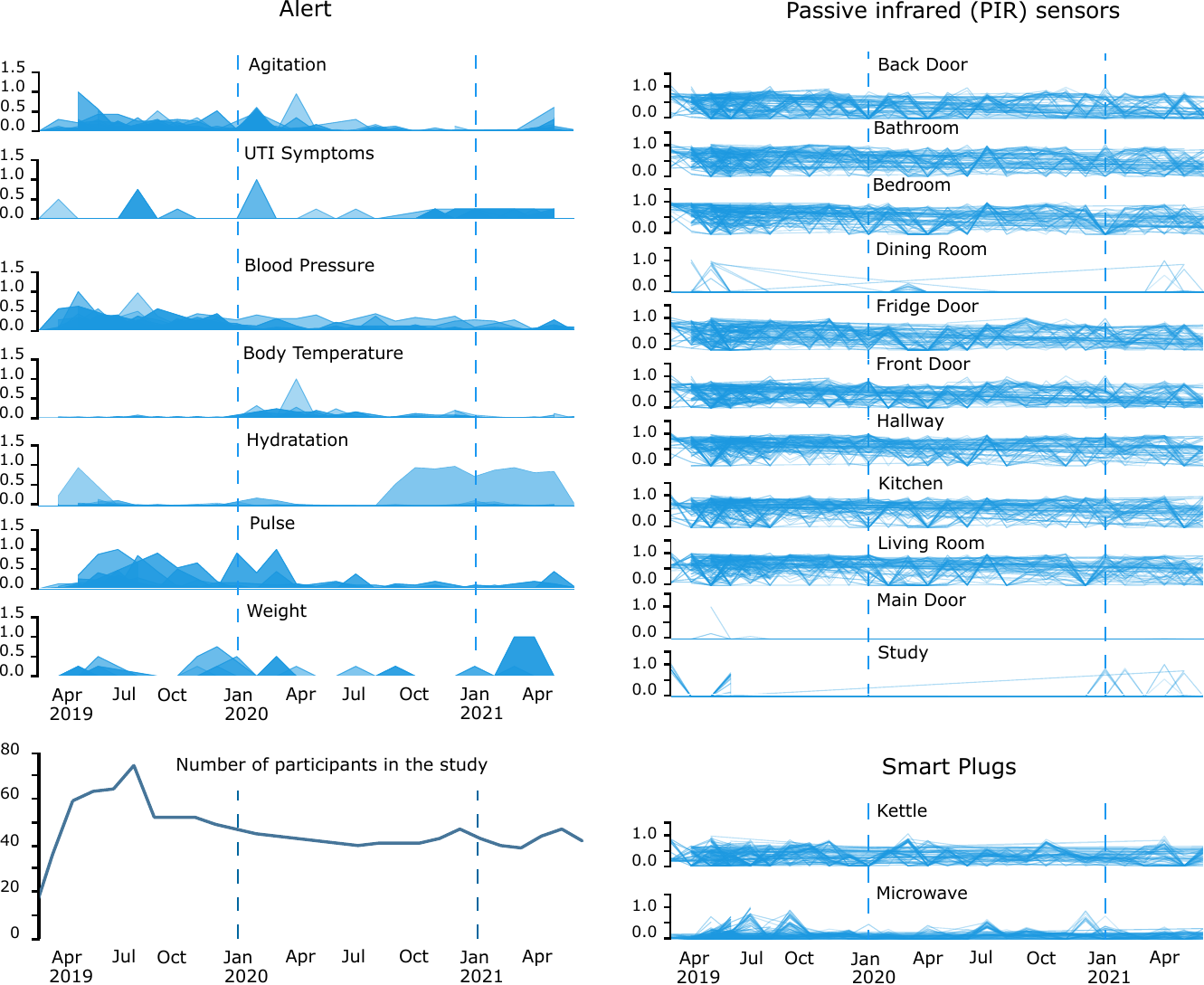}
    \caption{The normalised distribution of the raised flagged events, movements and smart plugs activation through two years period. 
    Agitations episodes consist of a total $\sim$1400 episodes (true and false).
    }
    \label{fig:flag_count}
\end{figure}

The proposed model performs a risk analysis in six-hour windows to assess the risk of agitation in PLWD. 
Figure~\ref{fig:flag_count} shows the normalised distribution of raised flags by the total amount of participants in the study (Total agitation events is equal to $\sim$1400 events). 
Figure~\ref{fig:flag_count} also demonstrates the total activated passive infrared sensors and smart plugs. 
In this study, we have only included the labelled and validated episodes of agitation. 
However, as shown in Figure~\ref{fig:flag_count}, the dataset includes verified episodes of other adverse health conditions such as Urinary Tract Infections (UTIs). 
%
\begin{figure}
    \centering
    \includegraphics{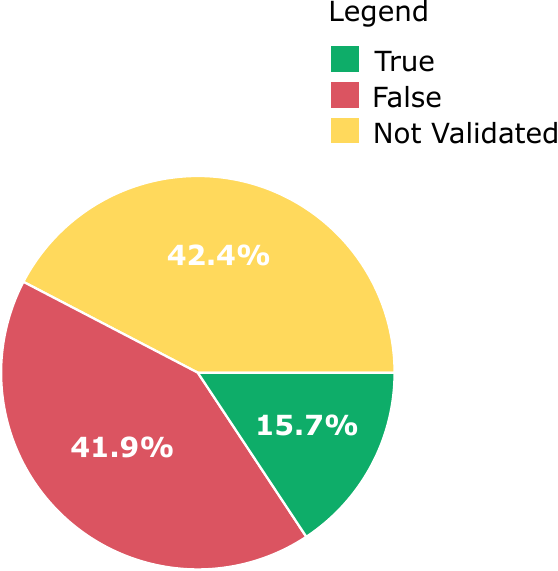}
    \caption{Distribution of the labels (True, False and Non-validated) among raised alerts associated with agitation. 
    The non-validated episodes are those for which our earlier triage methods raised an alert, but the monitoring team could not contact and verify the episode due to not having access to PLWD or their carers or clinical decisions to contact the participants only on occasions that the raw observation and measurement also demonstrated abnormality to reduce the burden on the participants during the data collection phase.}
    \label{fig:label_count}
\end{figure}
The distribution of labels among the three possible classes is depicted in Figure \ref{fig:label_count} (not validated, false, and true). 
Due to the sporadic occurrences of agitation episodes, the dataset used in the experiments contains more false samples than true ones. 
Consequently, the dataset is imbalanced with True class making up just 15.70\% of the total number of labels. 
In the current work, the non-validated labels are dropped, however, a weak learning method can be introduced in the future to investigate the improvement of the model's robustness.

To limit the amount of noisy and unreliable labels, alerts that took more than three days to be verified by the monitoring team were also removed. 
Furthermore, the original set of labels is expanded by adding extended labels created by slicing the time window of the existing labels. 
For example, if a true validated agitation episode was created at 6 PM, five additional labels were created to cover each of the hours in the time windows as a separate agitation label. 
In total, six entries were included for each validated episode. 
The final dataset used for the experiments consists of data from 46 participants' homes, 600 true labels and 1554 false labels. 

\subsection{Model Design}
\label{sec:lstm}
Long-Short Term Memory (LSTM) Networks~\cite{hochreiter1997long} are a special type of Recurrent Neural Networks (RNN), which are able to learn long-term dependencies using gates that regulate the flow of information.
LSTM networks have been implemented in the prediction and forecasting of time-series data in several healthcare applications~\cite{lu2018deep, pham2017predicting}. 

\begin{figure}
    \centering
    \includegraphics{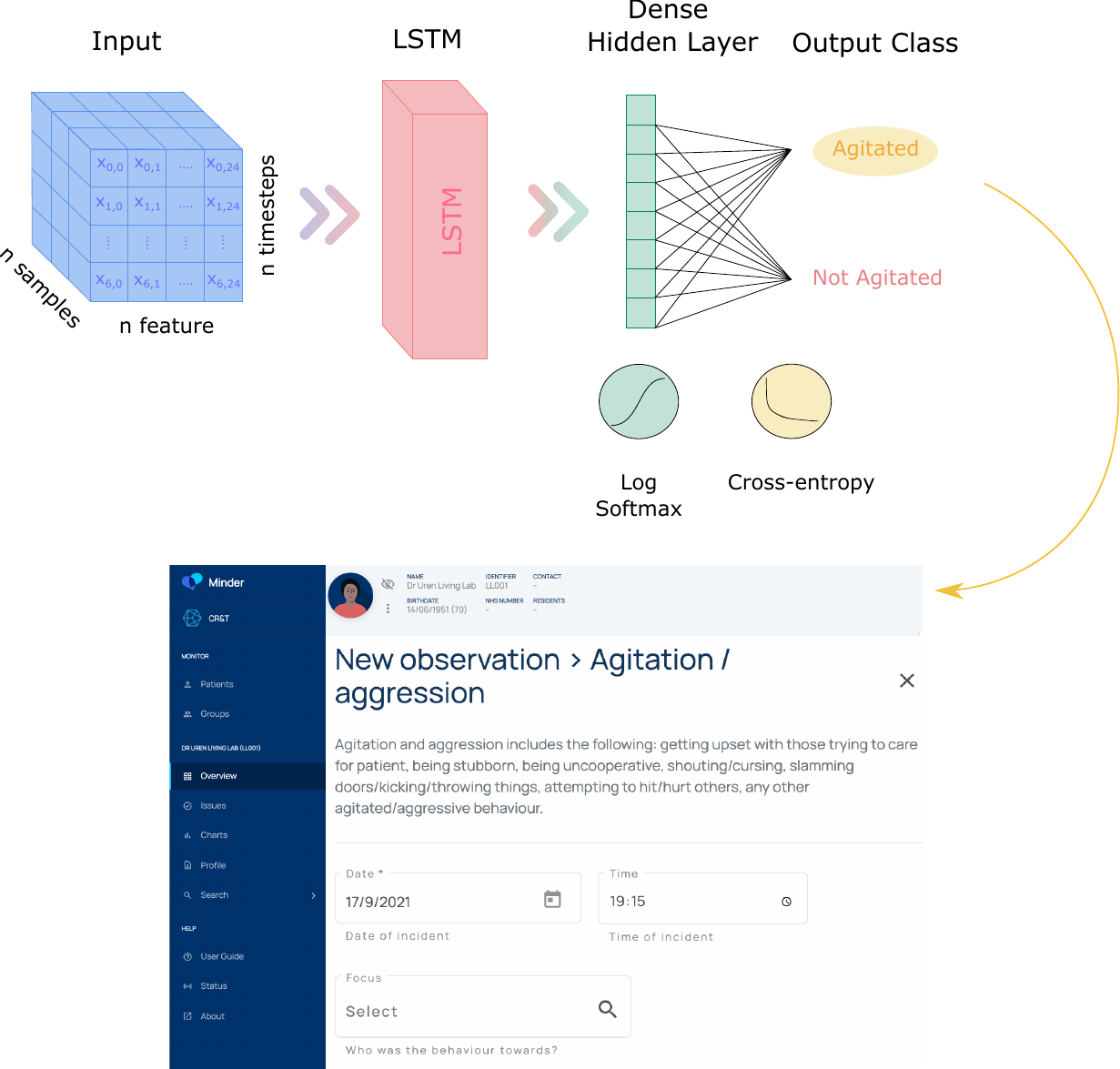}
    \caption{Architecture of the implemented LSTM model. 
    The 3D input data (n\_timestamps $*$ n\_features $*$ n\_samples) are feed into an LSTM model followed by a Dropout Layer.
    The resulting data are then analysed by one hidden layer with $log softmax$ activation and cross-entropy loss, which then predicts the class of each sample. 
    Adam is used as the optimiser. 
    We plan to integrate the model into the Minder digital platform; so an alert will be raised when an agitation episode is detected.
    }
    \label{fig:model_architecture}
\end{figure}

In this work, an LSTM network is implemented to analyse the risk of agitation episodes in PLWD.
The first layer of the sequential model consists of the LSTM, $tanh$ activation and $sigmoid$ continuous activation. 
This is followed by a Dropout layer, set to a 0.4 rate that helps prevent overfitting. 
A $log softmax$ with two neurons (the two possible values True or False for agitation) is implemented at the output layer.
Categorical cross-entropy is used as loss and Adam as optimiser. 
A $softmax$ output layer was also investigated, but it was found that the recall was not comparable in respect to $log softmax$. 
Using $log softmax$ allows better optimisation of the gradients and improve the sensitivity of the model. 
In the experiments, following the initial labelling approach that used 6 hours intervals, six one-hour entries are created for each existing validated sample. 
Each data point contains 6 time points and 24 features.
Figure \ref{fig:model_architecture} shows the architecture of the implemented model.
The same architecture is also adapted by introducing a Bidirectional LSTM (B-LSTM) \cite{schuster1997bidirectional} instead of the basic LSTM to investigate model performance by using information from the preceding and proceeding time windows. 
The best combination of parameters for LSTM and B-LSTM architectures are investigated in Section \ref{sec:experiments_setup}.
\section{Experiments Setup and Discussion}
\label{sec:experiments_setup}
To validate the model, the LSTM and B-LSTM architectures introduced in Section \ref{sec:lstm} are tested on the dataset created in Section \ref{sec:preprocessing}.
The optimal values for batch size, epochs, number of neurons, the introduction of a layer normalisation after the LSTM, and the usage of class weights to balance the classes are investigated through a grid search and shown below.
The dataset is divided into 70\% for training, 20\% for validation and 10\% for test. 
Each set was chosen in such a way that the data of each participant did not overlap across the training, validation, and test sets. 
A total of 46 participants which had sufficient data to be included in the experiment were extracted from the total dataset. 
The other participants' data were not considered in this study as the data had more than 50\% missing values, participants had left the study or they did not have the complete set of in-home devices that were required for this study. 
The 46 participants were randomly divided to 33 for training, 8 for validation and 5 for testing in each round of experiments. 
Because the splitting is done by participant ID, some participants may have less data or not true labels for agitation.
To better investigate the variability of the data, the group of participants used to partition the dataset into training, validation, and testing is randomly shuffled and different participants are used at each repetition.
When performing the shuffle, a random seed equal to the number of the current repetition is employed to maintain repeatability in the study.
To investigate the best combination of parameters to train the LSTM architectures and the variability of the dataset, a total of 10 repetitions are performed for each combination of parameters (1,152 possible combinations) for a total of 11,520 repetitions. 
Class weights are calculated and added while fitting the model to balance the disequilibrium among the classes in the dataset and to weight the loss function during training of the model. 
The models were trained and tested with Tensorflow \cite{abadi2016tensorflow} on a computer with Windows OS, AMD Ryzen 9 5900 processor, using 32GB RAM and GPU GeForce RTX 3070 6GB.

\begin{figure}
    \centering
    \includegraphics{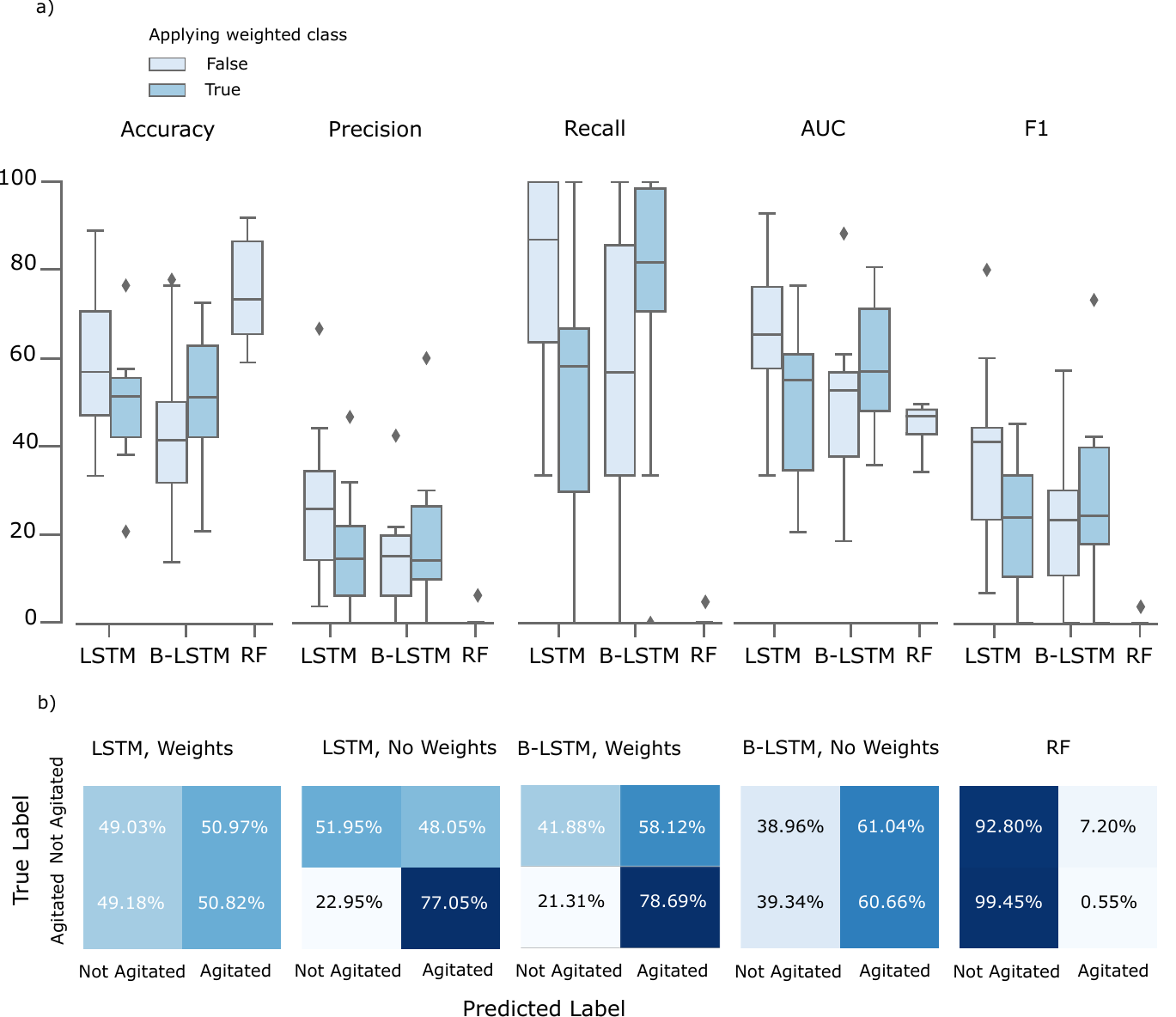}
    \caption{a) Results of analysis for the detection of agitation episodes in people affected by dementia. 
    The best combination of parameters was 32 for batch size, 300 for epochs and 200 for the number of neurons for LSTM with no class weights, which achieved an average of $\sim$60\% of classification accuracy and $\sim$85\% recall score. 
    b) Confusion Matrices for the proposed models. The LSTM with no class weights is able to correctly identify $\sim$77\% of agitation episodes}
    \label{fig:results}
\end{figure}

Youden's J statistic is used to analyse the performance of the models and identify the best combination of parameters.
This metric takes into consideration both sensitivity and specificity and can be used to summarise the performance of a diagnostic test:
\begin{equation}
    J = sensitivity + specificity - 1
\end{equation}
Implementing grid search and the Youden's J Index, it is found that the best combination of parameters is equal to 32 for batch size, 300 for epochs and 200 for the number of neurons for LSTM with no class weights achieving a $\sim$0.32 Youden's Index while B-LSTM with class weights achieves $\sim$0.23 Youden's Index.
Furthermore, adding layer normalisation in the architecture didn't improve the results.
Figure \ref{fig:results} shows the results of the analysis when the LSTM and B-LSTM models are trained and tested using the optimal set of parameters in combination with the class weights.
The models are compared with a RF classifier with 100 trees trained and tested on the same dataset.
Accuracy, precision, recall, Area Under the Receiver Operating Characteristics curve (AUC), and F1-Score are used as metrics to evaluate the model.
The confusion matrix of the five compared models (LSTM and B-LSTM with and without class weights and RF) is also shown.
As deduced by the Youden's J Index, the model with the best balance between false positive and false negative is the LSTM with no class weights implemented.
Using this combination of parameters allows identifying 77.05\% of agitated episodes on an average of 10 runs with different splits of the training and test sets.
We note the $log softmax$ activation already considers the unbalanced dataset and penalises the model when it fails to predict the correct class.

\begin{figure}
    \centering
    \includegraphics{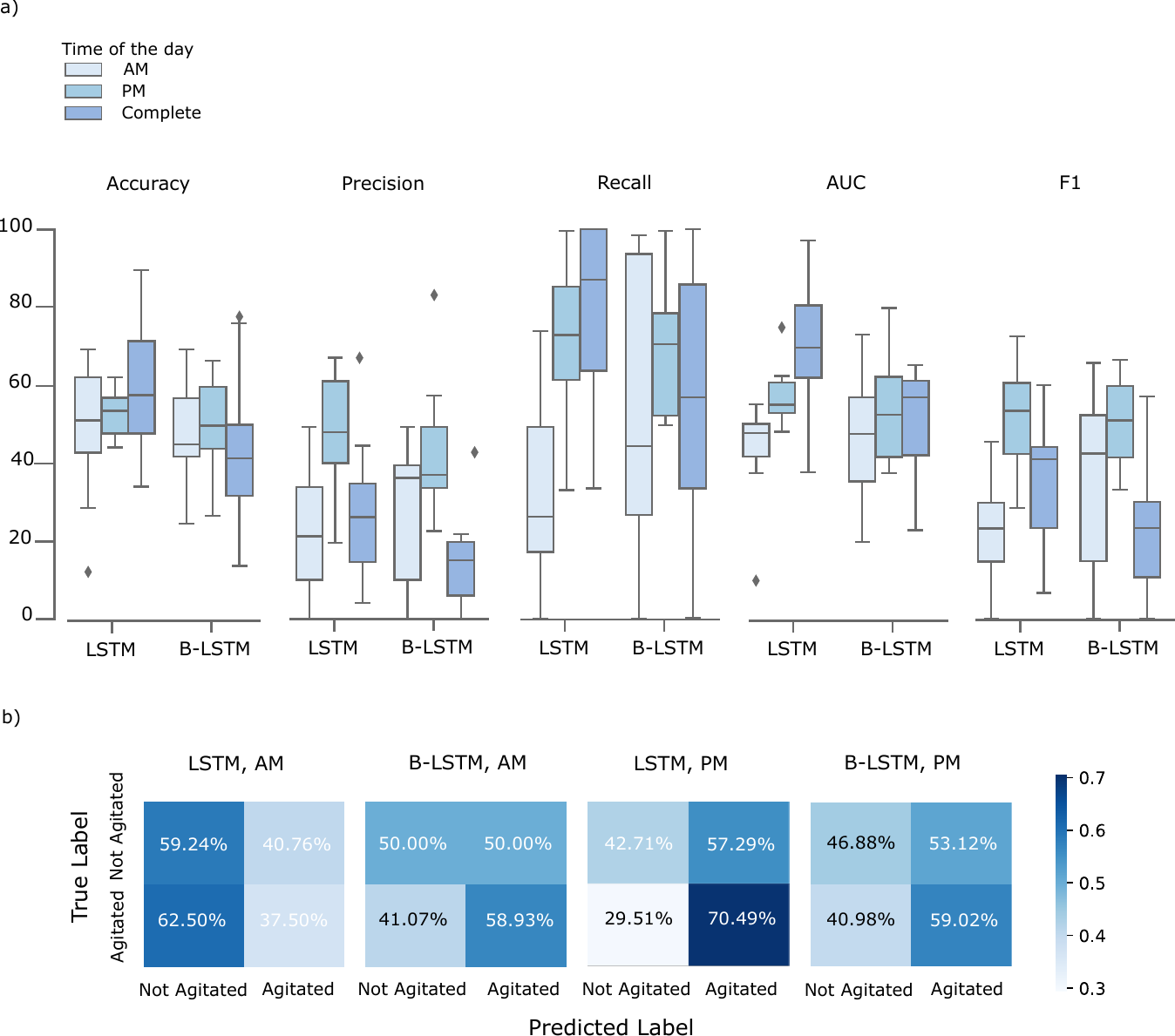}
    \caption{a) Results of the complete analysis for the detection of agitation episodes in people affected by dementia compared with two further datasets extracted from the complete data based on the time of the creation of the label: AM, which ranges from 00:00 to 12:00 and PM from 12:00 to 24:00 (00:00). 
    b) Confusion Matrices for the proposed models for AM and PM only experiments. 
    The LSTM with no class weights trained and tested on the PM dataset is able to identify $\sim$70\% of agitation episodes correctly.}
    \label{fig:time_comparison}
\end{figure}

A second experiment has been performed to investigate the possible effect of the time of the day on the agitation state.
Two datasets are created based on the hour of the day: AM, which goes from 00:00 to 12:00 and PM from 12:00 to 24:00 (00:00).
The two datasets are then used to train the two models described in \ref{sec:lstm} using the optimal set of parameters obtained via grid search and no class weight.
Figure \ref{fig:time_comparison}~a) shows the results and compares the two AM and PM datasets with the full dataset.
Using the complete dataset achieves higher results for both accuracy and recall considering the LSTM network, which was the best performing model in the previous experiment. 
The precision and F1-score are higher when using the PM only dataset.
This may be due to the less amount of not-agitated labels in the PM dataset.
Figure \ref{fig:time_comparison}~b) shows the confusion matrix for AM and PM analysis. 
LSTM trained and tested on PM dataset achieves the best results for the classification of agitation episodes with a total of 70.49\% correctly classified. 

In analysing behavioural changes in the home and to triage the response and interventions, False Negative outputs are considered more important than False Positive as it is important to detect as many agitation episodes as possible especially considering the human experts in the loop to investigate the alerts. 
Agitation episodes need to be accurately detected and not overlooked while if non-agitation events are mislabelled as agitation they could be investigated by the monitoring team before making any decision or taking any actions. 
However, a high level of misclassification will add a burden on the monitoring team in verifying them and could lead to unnecessary calls to PLWD and their carer when a false alert is raised. 
To address this issue, the monitoring team can investigate the underlying data and contact the participants during their routine daily/weekly calls and verify any existing alerts.  
It should be noted that the current study included agitation episodes for a period of $\sim$787 days, over two years.
Thus, a total of 148 false positive over that period equals an average of $\sim$6 calls per month.

\section{Limitations of the Study}
\label{sec:discussion}
We would like to highlight that this is a preliminary work on using in-home monitoring data to detect changes in behavioural patterns. 
Further research is required to improve collection and personalisation of in-home movement data and to increase the model robustness and generalisation. 

The data for this investigation was gathered in an uncontrolled, real-world environment. 
The sensors are placed in the homes of PLWD, the data collected is noisy and may contain information from other people in the home (e.g. family members, carers). 
A system to distinguish between the activation of sensors by various people will improve the robustness of the model. 
It should be also noted that a large part of the current data is unlabelled ($\sim$42.4\%). 
One direction for future work will be applying weak learning and semi-supervised models to utilise the unlabelled data. 
Furthermore, because the labelling is done by a clinical monitoring team and validated by contacting PLWD and their carers, in some cases it may contain noise and inaccuracies. Future work will focus on allowing the monitoring team to add confidence scores to the labels and to develop models that learn from noisy labelled data. 

\section{Conclusion and Future Work}
\label{sec:conclusion}
Agitation is one of the symptoms of dementia that can significantly impact ADL in PLWD. 
Detecting agitation episodes can help clinicians and care teams to provide more personalised and timely interventions by analysing the changes of behavioural patterns. 
In this work, we presented a preliminary model to analyse the risk of agitation in PLWD. 
The work compares an LSTM with a B-LSTM to analyse the continuous in-home monitoring data.

To test the efficacy of the proposed model, two experiments were devised: detect agitation episodes over the full dataset and compare agitation episodes occurring in the first and second halves of the day. 
An LSTM architecture with a $log softmax$ activation at the final layer and cross-entropy loss was implemented.
The top performing classifier (LSTM) correctly identified $\sim$75\% of the total cases of agitation in the test set over 10 repetitions in which the training, validation and test sets were shuffled at each run. 
A second experiment has been performed for testing the data based on the time of raised agitation episodes. 
LSTM architecture, with the same optimal parameters as the first experiment, achieved the best results when trained and tested on data from the PM dataset (ranging from 12:00 to 24:00). 

In future investigations, the input data will be extended to include more digital markers including sleep data and environmental information such as light and temperature which have relation to behavioural patterns \cite{gehrman2003sleep}. 
The ability of the model to generalise to various settings and diverse groups of participants will be further examined by training and testing the model on data with further data collection from a larger number of participants in our study. 
The use of Minder Digital Platform and low-cost in-home monitoring data with a human in the loop workflow design allows us to collect continuous data and test and validate the model in clinical care settings. 
As the data collection study progresses and as more samples and labelled data are collected via validation of the results and predictions of the current model, further work will be carried out to improve the efficacy of the model and integrate the model into the digital platform. 
Using a robust model to analyse the changes to behavioural patterns will provide continuous data and actionable insights to clinicians and support teams in dementia care. 

\bibliographystyle{abbrvnat}
\bibliography{references}

\end{document}